\documentclass[letterpaper, 10 pt, conference]{ieeeconf}  

\IEEEoverridecommandlockouts                              

\usepackage{adjustbox}
\usepackage{multirow}
\usepackage{caption}
\usepackage{graphicx}
\usepackage{booktabs} 
\usepackage{float} 
\usepackage{subcaption}

\title{\LARGE \bf
Automatic Cough Analysis for Non-Small Cell Lung Cancer Detection
}

\author{Chiara~Giangregorio$^{1}$, Cristina~Maria~Licciardello$^{1}$, Vanja Miskovic$^{1,2}$, Leonardo~Provenzano$^{1,2}$,  \\Alessandra Laura Giulia Pedrocchi$^{1}$, Andra Diana Dumitrascu$^{2}$, Arsela~Prelaj$^{2}$,\\Marina Chiara Garassino$^{3}$, Emilia~Ambrosini$^{1}$*, Simona~Ferrante$^{1,4}$*
\thanks{This work was supported by the European project I3LUNG (Grant Agreement: 101057695).}
\thanks{$^{1}$C. Giangregorio (corresponding author: chiara.giangregorio@polimi.it), C.M. Licciardello, V. Miskovic, L. Provenzano, A. Pedrocchi, S. Ferrante and E. Ambrosini are with Department of Electronics, Information and Bioengineering, Politecnico di Milano, Piazza Leonardo da Vinci 32, Milan, Italy}%
\thanks{$^{2}$ V.Miskovic, L. Provenzano, A.D. Dumitrascu, A. Prelaj are with Fondazione IRCCS Istituto Nazionale dei Tumori di Milano, Milan, Italy}
\thanks{$^{3}$ M.C. Garassino is with Department of Medicine, Section of Hematology/Oncology, University of Chicago, Chicago, IL, USA}
\thanks{$^{4}$ S. Ferrante is with the LEARNLab, IRCCS Istituto Neurologico Carlo Besta, Milan, Italy}
\thanks{*E. Ambrosini and S. Ferrante contributed equally to this work.}
}

\begin{document}

\maketitle
\thispagestyle{empty}
\pagestyle{empty}

\begin{abstract}
Early detection of non-small cell lung cancer (NSCLC) is critical for improving patient outcomes, and novel approaches are needed to facilitate early diagnosis. In this study, we explore the use of automatic cough analysis as a pre-screening tool for distinguishing between NSCLC patients and healthy controls. Cough audio recordings were prospectively acquired from a total of 227 subjects, divided into NSCLC patients and healthy controls. The recordings were analyzed using machine learning techniques, such as support vector machine (SVM) and XGBoost, as well as deep learning approaches, specifically convolutional neural networks (CNN) and transfer learning with VGG16. To enhance the interpretability of the machine learning model, we utilized Shapley Additive Explanations (SHAP). The fairness of the models across demographic groups was assessed by comparing the performance of the best model across different age groups (\(\leq\)58y and \(>\)58y) and gender using the equalized odds difference on the test set.  The results demonstrate that CNN achieves the best performance, with an accuracy of 0.83 on the test set. Nevertheless, SVM achieves slightly lower performances (accuracy of 0.76 in validation and 0.78 in the test set), making it suitable in contexts with low computational power. The use of SHAP for SVM interpretation further enhances model transparency, making it more trustworthy for clinical applications. Fairness analysis shows slightly higher disparity across age (0.15) than gender (0.09) on the test set. Therefore, to strengthen our findings' reliability, a larger, more diverse, and unbiased dataset is needed—particularly including individuals at risk of NSCLC and those in early disease stages.
\newline
\indent \textit{Clinical relevance}— This study highlights the promise of integrating automatic cough analysis with machine learning techniques for improving lung cancer screening methods.
\end{abstract}

\section{Introduction}
Worldwide, lung cancer is one of the most frequently diagnosed cancers and the leading cause of cancer-related death \cite{Barta-2019}. It is classified into two main types: non-small cell lung cancer (NSCLC) and small cell lung cancer (SCLC), based on their histopathological characteristics and cellular morphology under microscopic examination \cite{NSCLC-SCLC}. About 85\% of the cases are classified as NSCLC, which usually progresses more slowly than SCLC \cite{NSCLC-epi}. However, by the time NSCLC is diagnosed, 40\% of cases have already spread beyond the lungs~\cite{NSCLC-prognosis}. 
Detecting NSCLC at an early stage provides the most promising prognosis. However, diagnosing NSCLC remains challenging, as its symptoms often mimic those of more common diseases or the long-term effects of smoking \cite{BALATA20191513}. Low-dose chest CT (LDCT) has been shown to reduce mortality by enabling early detection; however, its widespread implementation requires screening a massive population, exposing individuals to radiation from X-rays. Additionally, traditional imaging methods can be costly and often remain inaccessible for routine screening. Consequently, 80\% of NSCLC cases are diagnosed at advanced stages, significantly complicating treatment efforts and reducing survival rates \cite{Bareschino2011}. 
This highlights the need for alternative approaches that are both scalable and affordable for early and accessible detection.\\
Cough, a common but often neglected symptom of lung cancer, offers an opportunity as a digital biomarker. While artificial intelligence (AI) has been successfully applied to classify cough in conditions such as COVID-19, pertussis, bronchitis, or pneumonia~\cite{CoughReview}, its potential for detecting NSCLC remains largely underexplored.  Indeed, the potential integration of AI-driven cough analysis into everyday devices like smartphones would offer a real-time, cost-effective, and scalable solution for respiratory disease screening~\cite{Ghrabli2024, Barata2022, Wanasinghe_Bandara_Madusanka_Meedeniya_Bandara_de2024}. During the COVID-19 pandemic, AI-based cough classification gained significant attention, with researchers leveraging machine learning (ML) models to enhance the speed and efficiency of screening efforts~\cite{Tena2021Automated, Vijayakumar2020Low, Manshouri2021Identifying, Andreu-Perez2021A, Imran2020AI4COVID-19:, Laguarta2020COVID-19}. 
One significant study presented by Pahar et al. \cite{PAHAR2021104572} proposed a machine learning-based COVID-19 cough classifier that could discriminate between COVID-19 positive coughs, COVID-19 negative coughs, and healthy coughs, all recorded on smartphones, achieving an area under the ROC curve (AUC) of 0.98.\\
Moreover, research carried out by Bales et al.~\cite{Bales2020} demonstrated how machine learning models and convolutional neural networks (CNNs) can effectively detect various respiratory conditions, such as bronchitis and pertussis, based on cough sounds. The study achieved over 89\% accuracy in both cough detection and disease classification, showcasing the versatility and potential of ML in diagnosing a range of respiratory conditions.\\
Similarly, Google researchers presented their Health Acoustic Representations (HeAR) system, which leverages large-scale self-supervised learning (SSL) to process health-related acoustic signals, such as coughs and breaths. Trained on a vast dataset of over 313 million two-second audio clips, HeAR aims to create a general-purpose audio encoder that excels across a range of health acoustic tasks, including cough classification and lung function analysis, such as spirometry. This research highlights the potential of SSL to enhance generalization in health acoustic tasks, offering potential scalable and accessible solutions for healthcare~\cite{Baur2024HeARH}.\\
In this context, the primary objective of this study is to investigate the feasibility of employing cough as a digital biomarker to differentiate between healthy individuals and NSCLC patients. Specifically, the study explores the potential of ML and deep learning (DL) techniques to classify forced cough sounds, offering a non-invasive approach to lung cancer detection.\\ 
By applying AI techniques, this work seeks to contribute to understanding how computational methods can support personalized medicine, trying to identify potential improvements in the detection and management of lung cancer through non-invasive methods.
\section{Methods}
\subsection{Data Collection and Pre-processing}
Forced cough recordings were collected from two groups: healthy control subjects (Healthy group) and patients with stage IIIB-IV NSCLC undergoing immunotherapy (Cancer group). For each participant, a single cough recording, along with their age and sex and smoking habits, was recorded. Participants signed written informed consent prior to inclusion in the study. The study was conducted as part of the EU-horizon project I3LUNG (NCT05537922) and was approved by the Ethical Committee of Fondazione IRCCS Istituto Nazionale dei Tumori di Milano (approval number 147/22, approved on 04/11/2024).\\
Audio recordings were acquired using a mobile app or a web app, with a sampling frequency of 16 kHz, and were stored on an Amazon Web Server. Pre-processing steps were implemented to address potential inconsistencies in audio quality due to varying recording devices and environmental conditions, including loudness normalization, low-pass filtering, and downsampling to 12kHz. After preprocessing, segmentation was performed to isolate meaningful cough events based on the cough segmentation algorithm developed by Orlandic et al.~\cite{Orlandic2020The}.  The recordings were then divided into 90\%, 10\% for training, and testing, respectively. 
The test set was used exclusively to evaluate the performance of the final models.\\
Mann-Whitney test for independent samples was applied to compare the two groups in terms of age. A Pearson Chi-squared test was instead used for biological sex and smoking habits.\\
Subsequently, two parallel feature extraction and analysis pipelines were carried out as inputs for classical machine learning and deep and transfer learning models.
\subsection{Machine Learning approach}
\subsubsection{Feature extraction}
A total of 39 acoustic features, described in Table~\ref{tab:features}, were computed in the time and frequency domain with Python libraries.\\
Non-parametric Mann-Whitney test was performed on the acoustic features to assess whether there were significant differences between the two groups.
\begin{table}[]
\caption{Description of extracted acoustic features}
\begin{adjustbox}{width=\linewidth}
\label{tab:features}
\begin{tabular}{|p{10em}|p{12em}|}
\hline
\textbf{Feature} & \textbf{Description}\\
\hline
Zero Crossing Rate (ZCR) \cite{MFCC-CF-ZCR, Spectral} & Measures how many times the waveform crosses the zero axis.\\
\hline
Root Mean Square Power & Measures root mean square power of the signal.\\ \hline
Dominant Frequency \cite{Spectral}& Frequency with the highest power.\\ \hline
Spectral centroid \cite{Spectral}& Center of mass of the spectrum\\ \hline
Spectral rolloff \cite{Spectral}& Frequency below which a specified percentage of the total spectral energy\\ \hline
Spectral spread \cite{Spectral}&  Standard deviation around the spectral centroid\\ \hline
Spectract skewness \cite{Spectral}& Measure of the asymmetry of the spectrum around the spectral centroid\\ \hline
Spectral kurtosis \cite{Spectral}& Measure of the flatness, or non-Gaussianity, of the spectrum around its centroid\\ \hline
Spectral bandwidth \cite{Spectral}&  Indication of how spread out the frequencies are in the spectrum\\ \hline
Spectral flatness \cite{Spectral}& Ratio of the geometric mean of the spectrum to the arithmetic mean of the spectrum\\ \hline
Spectral slope \cite{Spectral}& Amount of decrease of the spectrum.\\ \hline
Spectral decrease \cite{Spectral} &  Amount of decrease of the spectrum, in the lower frequencies.\\ \hline
Crest factor \cite{Orlandic2020The, MFCC-CF-ZCR} & Ratio of peak amplitude to root mean square value. \\ \hline
Mel-Frequency Cepstral Coefficients (0-16) \cite{MFCC-CF-ZCR} & Shape and energy of the vocal tract in different frequency bands, based on the human ear's perception of sound.\\ \hline
Cough Length & Duration of cough.\\ \hline
Power Spectral Density (PSD) (8) \cite{Orlandic2020The,PSD} & Power distribution over frequency bands (0-200, 300-425, 500-650, 950-1150, 1400-1800, 2300-2400, 2850-2950, 3800-3900). \\ \hline
\end{tabular}
\end{adjustbox}
\end{table}
To address the issue of collinearity, the Pearson correlation coefficient was calculated among all acoustic features. Then those with a correlation coefficient greater than 80\% were removed.
\subsubsection{Classification}
Support Vector Machines (SVM)~\cite{SVM}, Logistic Regression (LR), and XGBoost (XGB)~\cite{XGBoost} were used for model training. SVM excels in handling high-dimensional data and capturing non-linear relationships through kernel functions, while XGB, as a gradient-boosting ensemble method,  effectively models complex feature interactions. LR was included as a baseline model, due to its simplicity, interpretability and low computational cost.
Hyperparameter tuning was performed using a grid search and validating the results with a 5-fold cross-validation. The best hyperparameters and subsequently the best model were selected based on the accuracy obtained during cross-validation.\\
For the best machine learning model, model explainability was assessed with SHAP to understand feature contributions.
\subsection{Deep and Transfer Learning approach}
\subsubsection{Feature extraction}
Mel-spectrogram images were extracted with 128 mel bands from cough audio recordings, which visually represent the signal's frequency spectrum over time.
\subsubsection{Deep Learning}
A Convolutional Neural Network was implemented using the TensorFlow library. The architecture, shown in Table~\ref{tab:CNN}, consisted of four Conv2D layers, each followed by a max-pooling layer to reduce the spatial dimensions and capture hierarchical features from the input data. The first Conv2D layer had 32 filters, while the remaining three layers each use 128 filters, all with a kernel size of (3, 3). A small kernel was chosen to capture more detail from the spectrograms. The activation function used in each convolutional layer was the Rectified Linear Unit (ReLU). After the convolutional layers, the network included a flattening layer to convert the 2D feature maps into a 1D vector, followed by a dropout layer with a rate of 0.25 to reduce overfitting. A fully connected dense layer with 1024 units was used before the output layer, consisting of one unit with a sigmoid activation function for the binary classification task.
\begin{table}[h]
    \caption{Architecture of the CNN model}
        \begin{adjustbox}{width=\linewidth}
            \label{tab:CNN}
            \begin{tabular}{|c|c|c|}
                \hline
                    \textbf{Layer Type} & \textbf{Output Shape} & \textbf{Activation}\\\hline
                    Input & (224,224,3) &  -\\ \hline
                    Conv2D (32 filters) & (222,222,32)  & ReLU\\ \hline
                    MaxPooling2D & (111,111,32)  & -\\ \hline
                    Conv2D (128 filters) & (109,109,128)  & ReLU\\ \hline
                    MaxPooling2D & (54,54,128) & -\\ \hline
                    Conv2D (128 filters) & (52,52,128) & ReLU\\ \hline
                    MaxPooling2D & (26,26,128)  & -\\ \hline
                    Conv2D (128 filters) & (24,24,128) & ReLU\\ \hline
                    MaxPooling2D & (12,12,128) & -\\ \hline
                    Flatten & (18432) & -\\ \hline
                    Dropout (0.25) & (18432)  & - \\ \hline
                    Dense (1024 units) & (1024) & ReLU \\ \hline
                    Dense (1 unit) & (1)  & Sigmoid \\ \hline
        \end{tabular}
    \end{adjustbox}
\end{table}
\subsubsection{Transfer Learning}
Transfer learning (TL) was evaluated by exploiting the pre-trained VGG16 model~\cite{VGG16}. The model was loaded with ImageNet weights, excluding the top classification layers, to allow for customization for our binary classification task. The pre-trained layers in the base VGG16 model capture rich, general-purpose features from the input images, which should help improve model performance on a new, smaller dataset.
All layers of the VGG16 base model were frozen, meaning their weights were not updated during training. This approach retains the features learned from ImageNet while reducing the risk of overfitting on the new task and speeding up training. On top of the VGG16 model, a custom sequential model was added, including a flattening layer, a fully connected dense layer with 1024 units, and ReLU activation, followed by a dropout layer with a rate of 0.3 to prevent overfitting. The final layer was dense with a single unit and a sigmoid activation function.\\
\par Both the DL and TL models were compiled using the Adam optimizer with a learning rate of 0.001 and binary cross-entropy as the loss function. Accuracy was chosen as the evaluation metric. Early stopping was applied with patience of 15 epochs to avoid overfitting, monitoring the validation loss and restoring the model weights from the epoch with the best validation loss.\\
To evaluate the DL and TL models, a 5-fold cross-validation was conducted on the training set to account for variability in the training/validation splits and to ensure robustness of the performance estimates. After cross-validation, a final model was trained on the whole training set, by splitting it into approximately 70\% for training and 30\% for validation, while still keeping the test set untouched for final evaluation.
\subsection{Fairness evaluation}
To evaluate the fairness of the best-performing model, age was identified as a sensitive attribute, and the population was stratified into two groups based on the distribution of ages in the dataset: individuals younger than 58 years and those aged 58 years and older. Specifically, 58 corresponded to the third quartile of the healthy control group and approximately the 15th percentile of the cancer group. This segmentation allowed for an assessment of the model’s ability to predict cancer cases consistently across age groups while accounting for the underlying demographic disparities between healthy and affected populations. In addition to age, gender was also considered a sensitive attribute. In both cases, fairness was quantified using the Fairlearn library's Equalized Odds Difference (EOD) with the mean approach, measuring the average disparity in model performance between groups while reducing the influence of outliers, providing a more balanced and robust assessment of fairness. Indeed, EOD was chosen as a metric since it allows for a more stable measure, especially in contexts with class imbalance and low event rates, where ratio-based metrics can become unstable or misleading. By focusing on absolute differences in error rates between groups, EOD offers a precise and more reliable assessment of disparities, ensuring that fairness evaluations remain robust even when one class is underrepresented.
\section{RESULTS}
A total of 227 subjects were recruited. As reported in Table~\ref{tab:demographics}, the cancer group included 118 participants (median age = 71 years, interquartile range [IQR] = 62.25–75), while the healthy group included 109 participants (median age = 45 years, IQR = 28–58). The age distributions were significantly different (P-value \(<0.001\)). The proportion of males and females was 64\% and 36\% in the cancer group and 42\% and 58\% in the healthy group, respectively (P-value = 0.002).The proportion of participants who have ever smoked (current or former smokers) is significantly higher in the Cancer group (57.6\%) compared to the Healthy group (21.1\%), with a p-value of \(<\)0.0001. Conversely, a much larger proportion of participants in the Healthy group are never smokers (72.5\%) compared to the Cancer group (7.6\%). Additionally, the Cancer group has a higher percentage of participants with unreported smoking status (34.8\%) compared to the Healthy group (6.4\%).
\begin{table}[h]
    \caption{Demographic characteristics of Cancer and Healthy groups.}
    \label{tab:demographics}
    \begin{adjustbox}{width=\linewidth}
    \begin{tabular}{|p{6em}|c|c|c|}
    \hline
        \textbf{Characteristic} & \textbf{Cancer (n=118)} & \textbf{Healthy (n=109)} & \textbf{p-value} \\\hline
        \textbf{Age, median (IQR)} & 71 (62.25–75) & 45 (28–58) & \(<0.0001\) \\\hline
        \multirow{2}{6em}{\textbf{Sex, n (\%)}}
            & Male: 75 (63.6\%) & Male: 46 (42.2\%) & \multirow{2}{*}{0.002} \\
            & Female: 43 (36.4\%) & Female: 63 (57.8\%) & \\\hline
        \multirow{3}{6em}{\textbf{Smoking, n (\%)}}
            & Ever smoked: 68 (57.6\%) & Ever smoked: 23 (21.1\%) & \multirow{3}{*}{\(<0.0001\)} \\
            & Never: 9 (7.6\%) & Never: 79 (72.5\%) & \\
            & Not Given: 41 (34.8\%) & Not Given: 7 (6.4\%) & \\\hline
    \end{tabular}
    \end{adjustbox}
\end{table}

\subsection{Machine Learning}
\begin{figure*}
    \centering
    \includegraphics[width=\linewidth]{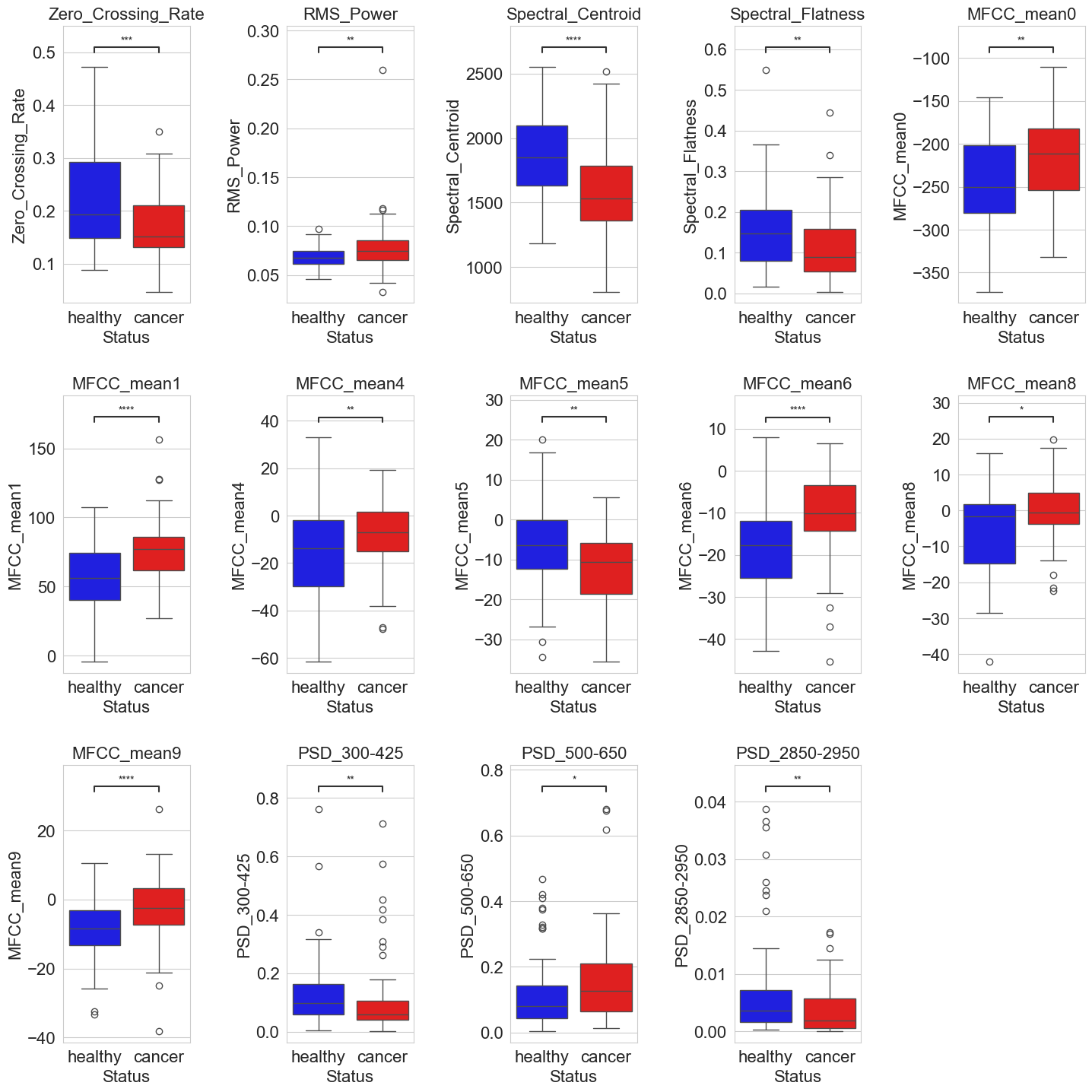}
    \caption{Feature distribution of the significantly different features among the Healthy and Cancer groups. Statistical significance levels are denoted by asterisks: * (p \(<\) 0.05), ** (p \(<\) 0.01), *** (p \(<\) 0.001).}
    \label{fig:boxplot}
\end{figure*}

After the correlation analysis on the train set, six features (i.e., spectral spread, skewness, kurtosis, bandwidth, slope and standard deviation) were removed due to collinearity.\\
Of the remaining 33 features, 14 significantly differed among the two groups and their distribution between the healthy and cancer groups is reported in Figure~\ref{fig:boxplot}. A higher zero crossing rate, spectral centroid, and flatness are observed in the healthy group compared to the cancer group. Furthermore, within the MFCC-related features, coefficients 0, 1, 4, 6, 8, and 9 exhibit a reduced frequency distribution in the healthy group, while MFCC 5 demonstrates elevated values. 
\subsubsection{Classification performances and model explainability}
\begin{figure}
    \centering
    \includegraphics[width=\linewidth]{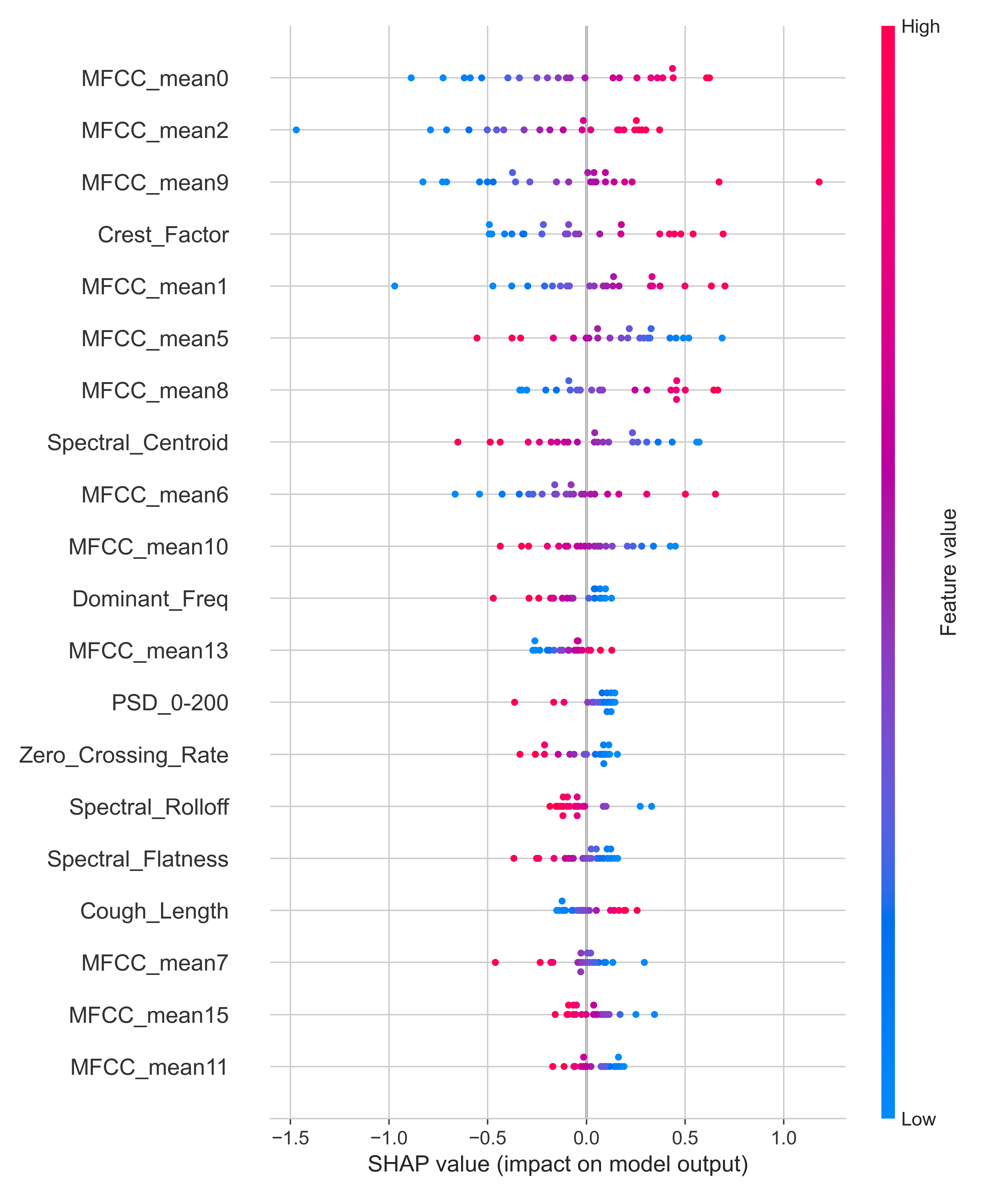}
    \caption{SHAP summary plot for SVM model}
    \label{fig:shap}
\end{figure}

Table~\ref{tab:acc_val} reports the accuracy obtained on the validation set for the ML models. SVM and XGB performed similarly on the validation set, both reaching 0.76. LR performed less effectively, with an accuracy of 0.74. The SHAP summary plot for SVM is presented in Figure~\ref{fig:shap}, which illustrates the impact of the features on the model's output for classifying test samples as healthy (left) or cancer (right). Each data point in the figure corresponds to a SHAP value for a specific feature in a test instance. The color of each data point indicates the feature's value, with red representing high values and blue representing low values. Features like MFCC\_mean0, MFCC\_mean2, MFCC\_mean9, and Crest\_Factor have the highest impact on the model's predictions, as indicated by their wider SHAP value ranges. In particular,  high values of these features appear to be more influential in predicting the cancer class, while their lower values contribute to healthy predictions.
\begin{table}[h]
    \centering
    \caption{Mean Accuracy (and standard deviation) of the models obtained during cross-validation: first three rows compare the performances of classical machine learning models; last two rows of the table compare the performances of deep and transfer learning models.}
    \begin{adjustbox}{width=.6\linewidth}
    \begin{tabular}{|c|c|}
    \hline
        \textbf{Model} & \textbf{Accuracy (validation)} \\ \hline
        \textbf{SVM} & \textbf{0.76} (0.05) \\ 
        XGBoost & 0.76 (0.07) \\ 
        LR & 0.74 (0.03) \\ \hline
        \textbf{CNN} & \textbf{0.78} (0.04) \\ 
        VGG16 & 0.77 (0.06) \\ \hline
    \end{tabular}
    \label{tab:acc_val}
    \end{adjustbox}
\end{table}

\subsection{Deep and Transfer Learning}
Table~\ref{tab:acc_val} reports the accuracy obtained during cross-validation for the deep (CNN), and transfer (VGG16) learning models.
The results demonstrate that the CNN model achieved a higher validation accuracy (0.78), while VGG16 obtained a slightly lower performance with an accuracy of 0.77.
\subsection{Performance Comparison on Test Set}
Additional metrics evaluated on the held-out test set for the best deep learning model (CNN) and best classical machine learning model (SVM) are reported in Table~\ref{tab:model_perf} for comparison. 
CNN exhibited superior precision, recall, and F1-score performance, achieving a macro-average F1-score of 0.83, whereas SVM obtained a macro-average F1-score of 0.78, showing good but comparatively lower performance. 
The CNN demonstrated higher precision for the healthy class (0.89) and higher recall for the cancer class (0.92), i.e., a higher sensitivity, with a stronger ability to correctly identify cancer cases than the SVM, which achieved a recall of 0.83 for the cancer class. Both models achieved the same recall for the healthy class (0.73), i.e., similar specificity, suggesting similar ability to classify healthy individuals correctly.
\begin{table}[h]
    \centering
    \caption{Accuracy, precision, recall, and F1-score on the held-out test set for SVM and CNN models.}
    \begin{adjustbox}{width=\linewidth}
    \begin{tabular}{|c|c|c|c|c|}
    \hline
        \textbf{Model} & \textbf{Metric} & \textbf{Healthy class} & \textbf{Cancer class} & \textbf{Macro average} \\ \hline
        \multirow{4}{*}{SVM} & Accuracy & - & - & 0.78 \\ \cline{2-5}
        & Precision & 0.80 & 0.77 & 0.78 \\ \cline{2-5}
         & Recall & 0.73 & 0.83 & 0.78 \\ \cline{2-5}
         & F1-score & 0.76 & 0.80 & 0.78 \\ \hline
        \multirow{4}{*}{CNN} & Accuracy & - & - & 0.83 \\ \cline{2-5}
        & Precision & 0.89 & 0.80 & 0.84 \\ \cline{2-5}
         & Recall & 0.73 & 0.92 & 0.83 \\ \cline{2-5}
         & F1-score & 0.80 & 0.86 & 0.83 \\ \hline
    \end{tabular}
    \label{tab:model_perf}
    \end{adjustbox}
\end{table}
\subsection{Fairness evaluation}
The mean EOD across the two age groups (\(>\)58 years and \(\leq\)58 years) for the CNN model was 0.15. The True Positive Rate (TPR) was slightly lower for individuals aged  \(>\)58 years (0.91) compared to those aged \(\leq\)58 years (1.00). Conversely, the False Positive Rate (FPR) was higher for individuals aged \(\leq\)58 years (0.20), while it was 0 for those aged \(>\)58 years.\\
For gender fairness, the mean EOD between male and female participants was 0.09. The TPR was lower for females (0.86) compared to males (1.00), while the FPR was slightly lower for females (0.17) than for males (0.20).
\section{Discussion}
The primary objective of this study was to explore whether cough could serve as a biomarker for non-small cell lung cancer, focusing on determining if it is possible to discriminate between the coughs of healthy individuals and those with NSCLC. The research explored the application of machine learning, deep learning, and transfer learning models to classify healthy and cancerous coughs using a set of extracted acoustic and spectral features. Preliminary statistical analyses performed on the acoustic features yielded insights into the discriminative power of individual features, thereby establishing a foundation for the models to leverage these differences. \\
The findings underscore the efficacy of CNN in this classification task, likely due to its capacity to capture complex patterns in the data, which can be challenging for traditional machine learning models such as SVM and LR. By comparison, VGG16 model demonstrated slightly lower performance, suggesting that the application of transfer learning with VGG16 may not have been optimal for this particular dataset or task, likely requiring further refinement and tuning.  Moreover, CNN model achieved an almost perfect recall of 0.92 for the cancer class on the test set, ensuring the detection of the vast majority of cancer cases - a crucial aspect of medical diagnosis. 
Nevertheless, CNN demonstrates higher performance on the test set than during cross-validation, which requires further investigation. This could be attributed to potential differences in data distributions, where the validation set might have been more challenging or less representative of the overall data. Conversely, SVM exhibits stable performance across cross-validation and test set, reflecting its robustness against data distribution shifts. Therefore, although SVM’s performance is lower than CNN on the test set, it remains a viable option due to its ability to achieve high accuracy and recall, especially in scenarios with limited computational resources.
In addition, SVM offered valuable insights into model explainability through SHAP analysis, thereby enhancing their trustworthiness. Indeed, the summary plot revealed the discriminative power of some features, such as MFCC 0, 2 and MFCC 9, which show a clear distinction between the healthy and cancer classes. Notably, high values of these coefficients appear to be more influential in predicting the cancer class, while their lower values contribute to healthy predictions. This aligns with the observed feature distributions in the boxplots, further validating the importance of these features in the classification task. These results suggest that these features could be capturing subtle acoustic variations potentially linked to differences in airflow dynamics, vocal tract behavior, or mucus production. Nevertheless, further investigation is needed to understand better their role in distinguishing between healthy and cancer-related cough patterns.\\
To assess the fairness of the best-performing model (i.e., CNN), EOD was computed across age and gender. For age fairness,  the mean EOD between the two age groups (\(\leq 58\) and \(> 58\)) was 0.15, indicating a mild difference between the groups. While the ideal EOD would be 0,  this result suggests that the model performs relatively well across both age groups, with only a slight difference in performance.
For gender fairness, the mean EOD was 0.09, reflecting minimal disparity. The TPR was lower for females (0.86) compared to males (1.00), while the FPR was slightly lower for females (0.17) than for males (0.20). These findings suggest an opportunity to refine the model further, improving fairness while maintaining its effectiveness.\\
While the results of this study highlight the potential of AI-based cough analysis for NSCLC detection, several limitations must be acknowledged. A key limitation is the dataset size. With only 227 samples available for training, validation, and testing, this dataset is relatively small for deep-learning models like CNN, which require significantly more data to generalize effectively. Expanding the dataset is essential for improving model robustness, reliability, and overall performance. 
Moreover, the current sample may not fully reflect individuals at risk for lung cancer—those who typically undergo screening. Table \ref{tab:demographics} highlighted differing proportions of individuals who have ever smoked between the Cancer and Healthy groups, as well as a markedly higher proportion of never smokers in the Healthy group, underscoring the need to include more smokers in the Healthy group to improve the clinical relevance and real-world applicability of the model.
In addition, given that early detection is critical to improving patient outcomes, future research should focus on training models with data from patients in earlier stages of the disease to assess the ability of AI to detect NSCLC at its earliest stages.
Another aspect that requires attention is the significant age difference between the two groups, that may cause a bias in model performance. To address this, the equalized odds difference was computed to assess fairness. However, future studies should include more balanced classes in terms of age to further mitigate potential biases and ensure equitable model performance.\\
To date, no studies have employed automatic cough analysis specifically for lung cancer detection. However, several low-invasive methods have been explored for this purpose. For instance, Chen et al. achieved a sensitivity of 0.86 and a specificity of 0.69 in distinguishing lung cancer patients from healthy controls using volatile organic compounds (VOCs) in exhaled breath \cite{Chen-VOC,VOCs}. Additionally, electronic nose (E-nose) technology for breath analysis has shown promise, with reported sensitivity ranging from 0.71 to 0.96 and specificity between 0.33 and 1.00 \cite{E-nose}. While these methods leverage respiratory biomarkers, they require specialized equipment and may be affected by environmental factors.
In contrast, cough analysis offers a simpler and more accessible alternative that has been successfully applied to detecting respiratory diseases such as pneumonia and COVID-19. Building on this foundation, this study demonstrates its potential for lung cancer detection, achieving overall a good specificity (0.73) and higher sensitivity (0.92) than existing breath-based methods. These findings highlight the potential of cough as a valuable digital biomarker for NSCLC. By providing a cost-effective, non-invasive screening tool, automatic cough analysis could facilitate earlier interventions and ultimately improve patient outcomes.

\section*{ACKNOWLEDGMENT}
The authors wish to thank all the patients and volunteers who contributed with their cough recordings to this study, as well as those involved in data acquisition. Their support was essential to this research.

\end{document}